\documentclass[a4paper,11pt,onecolumn]{quantumarticle}
\pdfoutput=1
\usepackage[utf8]{inputenc}
\usepackage[english]{babel}
\usepackage[T1]{fontenc}
\usepackage{graphicx}% Include figure files
\usepackage{amsmath}
\usepackage{amssymb}
\usepackage{hyperref}
\usepackage{quantikz}
\usepackage{braket}
\usepackage{float}
\usepackage{subcaption}
\usepackage{tikz}
\usepackage{lipsum}
\usepackage[numbers,sort&compress]{natbib}

\begin{document}

\title{Optimization of the quantization of dense neural networks from an exact QUBO formulation}
\author{Sergio Muñiz Subiñas}
 \email{sergio.muniz@itcl.es}
\affiliation{Instituto Tecnológico de Castilla y León, Burgos, Spain}
\orcid{0009-0008-7590-0149}

\author{Manuel L. González}
 \email{manuel.gonzalez@itcl.es}
\affiliation{Instituto Tecnológico de Castilla y León, Burgos, Spain}

\author{Jorge Ruiz Gómez}
 \email{jorge.ruiz@itcl.es}
\affiliation{Instituto Tecnológico de Castilla y León, Burgos, Spain}
\orcid{0009-0006-1357-1461}

\author{Alejandro Mata Ali}
 \email{alejandro.mata@itcl.es}
\affiliation{Instituto Tecnológico de Castilla y León, Burgos, Spain}
\orcid{0009-0006-7289-8827}

\author{Jorge Martínez Martín}
 \email{jorge.martinez@itcl.es}
\affiliation{Instituto Tecnológico de Castilla y León, Burgos, Spain}
\orcid{0009-0004-9336-3165}

\author{Miguel Franco Hernando}
 \email{miguel.franco@itcl.es}
\affiliation{Instituto Tecnológico de Castilla y León, Burgos, Spain}
\orcid{0009-0009-9059-6832}

\author{Ángel Miguel García-Vico}
 \email{angelmiguelgarciavico@gmail.com}
\affiliation{Andalusian Research Institute in Data Science and Computational Intelligence (DaSCI), University of Jaén, 23071 Jaén, Spain}
\orcid{0000-0003-1583-2128}

\author{Javier Sedano}
 \email{javier.sedano@itcl.es}
\affiliation{Instituto Tecnológico de Castilla y León, Burgos, Spain}
\orcid{0000-0002-4191-8438}
\maketitle

\begin{abstract}
This work introduces a post-training quantization (PTQ) method for dense neural networks via a novel ADAROUND-based QUBO formulation. Using the Frobenius distance between the theoretical output and the dequantized output (before the activation function) as the objective, an explicit QUBO whose binary variables represent the rounding choice for each weight and bias is obtained. Additionally, by exploiting the structure of the coefficient QUBO matrix, the global problem can be exactly decomposed into $n$ independent subproblems of size $f+1$, which can be efficiently solved using some heuristics such as simulated annealing. The approach is evaluated on MNIST, Fashion-MNIST, EMNIST, and CIFAR-10 across integer precisions from int8 to int1 and compared with a round-to-nearest traditional quantization methodology.
\end{abstract}

%\tableofcontents
\section{Introduction}
In recent years, neural networks have achieved remarkable success in several applications, such as image classification, regression, or natural language processing. Given its high computational cost \cite{energy_impact_Ia}, it is common to make approximations to reduce their complexity and improve efficiency. An adopted method is quantization, which involves rounding the network’s coefficients from floating-point to integers, reducing the required stored memory of the model and accelerating the operations \cite{general_quantization, white_paper}. There are two main quantization techniques: quantization-aware training (QAT) \cite{QAT-1, QAT-2}, incorporating quantization during the training phase, and post-training quantization (PTQ) \cite{PTQ-1,PTQ-2, PTQ-3}. Usually, QAT techniques obtain better performance but they are computationally more expensive. Therefore, this work focuses on PTQ techniques, which offer a more practical and efficient alternative. One of the most common PTQ strategies is the round-to-nearest quantization, where each value is rounded to the closest integer. There are other alternatives that are harder to implement but yield better results. For example: KL-divergence calibration selects the optimal clipping range by minimizing the divergence between the original and quantized activation distributions, and it is adopted in TensorRT~\cite{quantization_kl_divergence_migacz}; FlexRound replaces the typical element-wise addition with element-wise division~\cite{PTQ-2}; ADAROUND \cite{adaround, PTQ-3}, which operates by selecting whether the weights should be rounded to the upper or lower integer. The rounding decision is formulated as a binary optimization problem, where a solution value of 1 indicates rounding up and a value of 0 indicates rounding down. Although PTQ quantization offers several benefits, in general, most of the available methods present some common disadvantages such as the computational cost, the requirement of calibration data,  sensitivity to hyperparameters or hardware restrictions.

Given its straightforward implementation in comparison to other methods and its capacity to minimize an objective function, ADAROUND is a popular PTQ strategy. With the objective of amplifying the state-of-the-art of the ADAROUND post-training quantization, this paper proposes a reformulation of the ADAROUND quantization problem. The formulation of this work directly minimizes the Frobenius distance between the theoretical and the de-quantized outputs before the activation, realizing that the structure of this problem is the same as a QUBO problem. The proposed methodology formulates and solves a QUBO whose solution provides a quantization that theoretically minimizes the Frobenius-norm error, preserving as much as possible the original network behavior. Another advantage of this method of quantization is that QUBO is a well-studied class of combinatorial optimization problems, and there are several exact and approximate solvers readily available. Furthermore, this paper proposes an exact decoupling of the general QUBO problem into QUBO subproblems due to the intrinsic structure of the problem. Briefly, the main contributions of this work are:
\begin{itemize}
    \item  A novel and explicit QUBO formulation of the ADAROUND quantization process, where each binary variable represents a rounding decision for each individual weight.
    \item An exact decomposition method that divides the global QUBO problem into independent QUBO subproblems.
    \item The development of a methodology that quantizes neuronal networks using ADAROUND and employs simulated annealing in order to solve the associated QUBO subproblems.

\end{itemize}

The remainder of this paper is organized as follows. Section~\ref{sec: related works} reviews the ADAROUND method and explains the basic notation of the QUBO formulation, quantization of variables, and neural networks. Section~\ref{sec: mathematical formulation} presents the reformulation of the ADAROUND method as a QUBO model, including the mathematical derivation of the cost function and the structure of the QUBO coefficient matrix. Section~\ref{sec: subproblems} introduces the methodology that divides the general QUBO problem into independent subproblems. Section~\ref{sec: implementation} details the implementation of the proposed method, including the use of simulated annealing as a QUBO solver. Section~\ref{sec: experimentation} shows the experimental results for MNIST, MNIST-FASHION, EMNIST, and CIFAR-10, analyzing the impact of quantization in different bit quantizations. Finally, Section~\ref{sec: conclusions} concludes the paper and outlines potential directions for future research.

\section{Related works and preliminaries}\label{sec: related works}
This section reviews the fundamental concepts that are used in the proposed method. It introduces the QUBO formulation, followed by a review of the ADAROUND quantization strategy. Additionally, it outlines the basic principles of variable quantization, including linear quantization and the de-quantization functions. 
\subsection{Quadratic Unconstrained Binary Optimization
(QUBO) formulation}
The quantization employing the ADAROUND methodology proposed in this paper requires solving a QUBO problem. Formally, a QUBO problem can be expressed as the maximization of a quadratic function over binary variables:
\begin{equation}
    \max_{\textbf{x} \in \{0,1\}^n} \quad \textbf{x}^T Q \textbf{x} + \textbf{c}^T \textbf{x} + k,
\end{equation}
where $Q \in \mathbb{R}^{n \times n}$ is a symmetric matrix representing the quadratic coefficients , $\textbf{c} \in \mathbb{R}^n$ is a vector, and $k \in \mathbb{R}$ is a constant. The QUBO formulation can be used to solve problems of different disciplines such as portfolio optimization~\cite{QUBO_portfolio} or logistics~\cite{QUBO_logistics}

Due to the computational complexity of the QUBO problems, which are NP-hard in general~\cite{np_hardness}, obtaining exact solutions is infeasible for large-scale instances. As a result, it is common to employ approximate techniques in order to obtain sub-optimal solutions.  Exact approaches typically rely on branch-and-bound~\cite{brach_and_bound} and branch-and-cut~\cite{branch_cut} techniques, which systematically explore the solution space while discarding suboptimal branches. Some commercial solvers that implement these techniques are Gurobi~\cite{gurobi2024} and CPLEX~\cite{cplex2024}. Among heuristic methods, the most prominent
are simulated annealing~\cite{simmulated_annealing_kirk},  tabu search~\cite{tabu_search_glover} or genetic algorithms~\cite{genetic_holland}.

\subsection{Related work of the ADAROUND quantization method}
ADAROUND is a PTQ technique that uses a binary variable $v$ to determine whether each weight of the neural network should be quantized to the lowest ($v=0$) or the highest integer ($v=1$). There are different approaches to selecting the rounding criterion. The original formulation proposed in \cite{adaround} approximates the effect of quantization using a second-order Taylor expansion of the expected change in the loss function:
\begin{equation}
    \begin{gathered}
        \mathbb{E}\big[\mathcal{L}(\mathbf{x}_{\text{in}},\mathbf{y}_\text{out},\mathbf{w}+\Delta{\mathbf{w}}) - \mathcal{L}(\mathbf{x}_{\text{in}},\mathbf{y}_\text{out},\mathbf{w})\big] \approx \\  
        \mathbb{E}\Big[\Delta{\mathbf{w}}^\top \nabla_\mathbf{w} \mathcal{L}(\mathbf{x}_{\text{in}},\mathbf{y}_\text{out},\mathbf{w}) + 
        \frac{1}{2} \Delta{\mathbf{w}}^\top \text{H}\big(\mathcal{L}(\mathbf{x}_{\text{in}},\mathbf{y}_\text{out},\mathbf{w})\big) \Delta{\mathbf{w}} \Big] = \\
        \Delta{\mathbf{w}}^\top \mathbf{g}^{(\mathbf{w})} + \frac{1}{2} \Delta{\mathbf{w}}^\top H^{(\mathbf{w})} \Delta{\mathbf{w}},
    \end{gathered}
\end{equation}
 $\mathcal{L}(\mathbf{x}_{\text{in}},\mathbf{y}_\text{out},\mathbf{w})$ denotes the error metric to be minimized, which depends on the input vector $\mathbf{x}_{\text{in}}$, the output vector $\mathbf{y}_\text{out}$ and the weights of the network $\mathbf{w}$. After quantization, the perturbed error metric is represented as $\mathcal{L}(\mathbf{x}_{\text{in}},\mathbf{y}_\text{out},\mathbf{w}+\Delta{\mathbf{w}})$, where $\Delta{\mathbf{w}}$ denotes the quantization perturbation in the weights. In this work, scalar quantities are denoted by regular lowercase letters (e.g. \( a \), \( w \)), vectors are represented by bold lowercase letters (e.g. \( \mathbf{x} \), \( \mathbf{y} \), \( \mathbf{w} \)), and matrices are denoted using uppercase letters (e.g. \( W \), \( H \)). Additionally, to account for the influence of the entire dataset $\{(\mathbf{x}_{\text{in}}^\alpha,\mathbf{y}_\text{out}^\alpha)\}_{\alpha=1,...,t}$, the expected value of the error is considered. Assuming the network is fully trained, the gradient of the loss function with respect to the weights vanishes. Consequently, the expected gradient is zero ${\mathbf{g}}^{(w)}=\mathbb{E}[\nabla_w\mathcal{L}(\mathbf{x}_{\text{in}},\mathbf{y}_\text{out},\mathbf{w})]=0$, since the error metric is at a local minimum in the
weight space. Therefore, the problem consists in minimizing the second-order term in the Taylor expansion:
\begin{equation}
    \arg \min_{\Delta{\mathbf{w}}}(\Delta{\mathbf{w}}^T{H^{(\mathbf{w})}}\Delta{\mathbf{w}}),
\end{equation}
where ${H^{(\mathbf{w})}} = \text{H}\big(\mathcal{L}(\mathbf{x}_{\text{in}},\mathbf{y}_\text{out},\mathbf{w})\big)$ denotes the Hessian matrix of the loss function with respect to the weights. This minimization problem can be formulated as a QUBO problem, since ${H^{(\mathbf{w})}}$ is a symmetric matrix and the perturbation $\Delta \mathbf{w}$ can be discretized and represented by a binary vector $\mathbf{v} \in \{0,1\}^n$, so the problem becomes
\begin{equation}
    \arg \min_{\mathbf{v}}(\mathbf{v}^T{H^{(\mathbf{w})}}{\mathbf{\mathbf{v}}}).
\end{equation}
This methodology requires a Taylor approximation of the loss function. In contrast, the present work offers an exact detailed derivation of an alternative formulation, based on the mathematical development of the Frobenius distance between the theoretical output and the de-quantized output, both evaluated before the activation function. This development leads to the explicit function to be minimized, which is also shown to be a Quadratic Unconstrained Binary Optimization (QUBO) problem.

\subsection{Mathematical formulation of quantization of variables}\label{subsec: quantization of variables}

Quantization of variables is a well-established technique and has been extensively studied in previous works \cite{quantization_ml, quantization_ml_2}. To support the mathematical development of the ADAROUND quantization method, it is necessary to first introduce some fundamental principles of linear quantization and de-quantized functions, following the formulations described in \cite{quantization_1, quantization_2}. Let
$\beta$ and $\alpha$ denote the maximum and minimum weight values of the set of numbers to quantize, and let $b$ represent the number of bits used for quantization. Linear quantization is a function $f:\mathbb{R}\rightarrow\mathbb{Z}$ that transforms the variable $a$ into its quantized version $\hat{a}$
\begin{equation}
\hat{a}=\text{clip}\left(\text{round}\left(\frac{a}{s}\right)+z\right),
\end{equation}
where the scale factor $s$ and zero-point $z$ are defined as:
\begin{equation}
   s=\frac{\beta-\alpha}{2^b-1}  \qquad  z=-\text{round}(\beta s)-2^{b-1}.
\end{equation}
Here, the operator $\text{round}(\cdot)$ is understood as a generic rounding function, which maps inputs with real values to integers. Its precise definition may vary depending on the quantization strategy. For example, the notation $\text{round}_{\text{tn}}(\cdot)$ refers to the classical round-to-nearest scheme, where the values are rounded to the closest integer, rounding up when the fractional part is greater than 0.5, and rounded down otherwise. The clip function constrains a value within an interval, replacing values below the minimum with the minimum and values above the maximum with the maximum \cite{quantization_2}.
The inverse transformation, known as dequantization, maps the quantized value $\hat{a}$ back to the original continuous domain
\begin{equation}
    \check{a}=s(\hat{a}-z).
\end{equation}
Typically $a\neq\check{a}$ as a result of rounding errors and clipping during quantization. In this manner, the round-to-nearest quantization and its dequantization of a variable can be defined as
\begin{equation}
    \hat{a}=\text{clip}\left(\text{round}_{\text{tn}}\left(\frac{a}{s}\right)+z\right),
\end{equation}
\begin{equation}
    \check{a}=s\cdot\text{round}_{\text{tn}}\left(\frac{a}{s}\right)=s\tilde{a}.
    \label{eq: rtn_dequantization}
\end{equation}
In contrast, the ADAROUND technique replaces the rounding operation with a learnable binary decision. Instead of rounding to the nearest integer, in the ADAROUND quantization $\text{round}_{\text{ADAR}}(a,s,v) = \left\lfloor{\frac{a}{s}}\right\rfloor+v$, the value is rounded either up or down by introducing a binary variable
$v \in \{0,1\}$, which indicates whether to round up ($v$=1) or down ($v$=0)
\begin{equation}
    \hat{a}=\text{clip}\left(\text{round}_{\text{ADAR}}(a,s,v)+z\right)=\text{clip}\left(\left\lfloor{\frac{a}{s}}\right\rfloor+v+z\right),
\end{equation}
\begin{equation}
    \check{a}=s\left(\text{round}_{\text{ADAR}}(a,s,v) \right)=s\left(\left\lfloor{\frac{a}{s}}\right \rfloor+v \right)=s(\underline{a}+v).
    \label{eq: adround_dequantization}
\end{equation}
This formulation allows the rounding direction to be optimized per weight element, enabling more accurate reconstructions of the original model output after quantization.

\section{QUBO-ADAROUND problem of a neural network layer}\label{sec: mathematical formulation}
Given a dense neural network layer defined as 
\begin{equation}
f^{(l)}(\textbf{x}^{(l)})=\sigma^{(l)}(W^{(l)}\mathbf{x}^{(l)}+\mathbf{b}^{(l)}) =    \sigma^{(l)}(\mathbf{y}^{(l)}) 
\end{equation}
where $\textbf{x}^{(l)}\in \mathbb{R}^n$ is the input vector of layer $l$, $W^{(l)}\in \mathbb{R}^{m \times n}$ is a weight matrix, $\mathbf{b}^{(l)}\in \mathbb{R}^m$ is the bias, $\sigma^{(l)}$ is a non-linear activation function applied element-wise and $W^{(l)}\mathbf{x}^{(l)}+\mathbf{b}^{(l)} = \mathbf{y}^{(l)}$ is the output before activation,
 the ADAROUND quantization problem consists in determining a set of binary variables that govern the rounding direction applied to the weight matrices $W^{(l)}$ and the bias vectors $\mathbf{b}^{(l)}$. This section presents a novel formulation of this problem. From this point onward, all notation refers to a single generic layer, and superscripts indicating the layer index are omitted for the sake of clarity. Specifically, for every layer, each element of $w_{ij}$ and $b_i$ is associated with a binary variable $v_{ij}$ and $v_i$, respectively, which indicates whether the corresponding value should be rounded up or down during quantization. The objective is to find the configuration $\mathbf{v}=(v_{11},...,v_{1f},...,v_{n1},...,v_{nf},v_1,...,v_n)$ that minimizes the Frobenius distance between the theoretical output $\mathbf{y}$ and the dequantized output $\check{\mathbf{y}}$, both before activation:
\begin{equation}
    \arg \min_{\mathbf{v}}\left[\left\lVert\mathbf{y}-\check{\mathbf{y}}(\mathbf{v})\right\rVert^2\right].
\end{equation}
To simplify the problem, each layer is treated independently instead of tackling the multilayer network as a whole. Minimizing the full network at once would introduce higher-order terms (third-order or higher) into the formulation. The procedure to follow consists of expressing the coefficients of $\mathbf{y}$ and $\check{\mathbf{y}}$ as functions of the coefficients of $\mathbf{x}, W, \mathbf{b}$ and $\mathbf{v}$. Then, the Frobenius distance between these two expressions is algebraically expanded and simplified in order to isolate the terms that depend on the quantization variables. By minimizing this expression, the optimal values of the binary vector $\mathbf{v}$ can be determined. For now, only a single vector $\mathbf{x}$ is considered, but in order to correctly quantize the neural network, it is necessary to average this process for a set $\{(\mathbf{x}_{\text{in}}^\alpha,\mathbf{y}_\text{out}^\alpha)\}_{\alpha=1,...,m}$, being $m$ the number of data considered.

As a first step, the coefficients of the theoretical pre-activation output for a given input are
\begin{equation}
    y_i=\sum_{j=1}^f w_{ij}x_j+b_i,
\end{equation}
where $j \in\{1,..,f\}$ and $i \in\{1,..,n\}$, $f$ being the size of $\mathbf{x}$ and $n$ the size of $\mathbf{y}$. Using the ADAROUND dequantized versions of $w_{ij}$ and $b_i$ (Eq. \ref{eq: rtn_dequantization}), and the round-to-nearest dequantization for $x_i$ (Eq. \ref{eq: adround_dequantization}):
\begin{equation}
\begin{gathered}
   \check{w_{ij}} =  s_w\left(\left\lfloor{\frac{w_{ij}}{s_w}}\right \rfloor+v_{ij} \right)=s_w\left(\underline{w_{ij}}+v_{ij}\right ),\\
    \check{b_i} =  s_b\left(\left\lfloor{\frac{b_i}{s_b}}\right \rfloor+v_i \right)=s_b\left(\underline{b_i}+v_i \right ),\\
    \check{x_i}=s_x\cdot\text{round}_{\text{tn}}\left(\frac{x_i}{s_x}\right)=s_x\tilde{x_i},
   \end{gathered}
   \label{eq: dequantization of variables}
\end{equation}
it is possible to define the dequantized version of $y_i$
\begin{equation}
\begin{gathered}
\check{y_i}(\mathbf{v})=\sum_{j=1}^f\check{w_{ij}}\check{x}_j+\check{b_i}= \\
=\sum_{j=1}^f  s_ws_x[(\underline{w_{ij}}+v_{ij})\tilde{x}_j] +s_b(\underline{b_i}+v_i)=\\
= \sum_{j=1}^f s_ws_x\underline{w_{ij}}\tilde{x}_j+s_b\underline{b_i} + \sum_{j=1}^f s_ws_xv_{ij}\tilde{x}_j+s_bv_i.
\end{gathered}
\label{eq: output dequantized}
\end{equation}
This expression is composed of terms independent of the quantization variables $\mathbf{v}$ and terms that depend linearly on them. The former can be precomputed, while the latter forms the basis of the optimization objective.
By expanding the Frobenius distance $\left\lVert\mathbf{y}-\check{\mathbf{y}}(\mathbf{v})\right\rVert^2$ and applying the necessary algebraic simplifications outlined step by step in Appendix \ref{apendix: mathematical_dev}, the resulting expression is:
\begin{equation}
    \left\lVert\mathbf{y}-\check{\mathbf{y}}(\mathbf{v})\right\rVert^2=(s_ws_x)^2\left ( \left\lVert \mathbf{y}-\underline{\mathbf{y}}'\right\rVert^2+\mathcal{Q}(\mathbf{v})\right),
\end{equation}
where $\left\lVert \mathbf{y}-\underline{\mathbf{y}}'\right\rVert^2$ represents a constant term that does not depend on $v_{ij}$ or $v_i$, and is also defined in Appendix \ref{apendix: mathematical_dev}. In contrast, the second term $\mathcal{Q}(\mathbf{v})$ captures all pairwise and individual contributions of the quantization variables and takes the following quadratic form:
\begin{equation} \label{B_term}
\begin{gathered}
\mathcal{Q}(\mathbf{v}) =\sum_{i=1}^n\sum_{j=1}^f\tilde{x}_j(\tilde{x}_j-2d_i)v_{ij}^2+\sum_{i=1}^n\sum_{j=1}^f\sum_{k=1, k{\neq}j}^f\tilde{x}_j\tilde{x}_kv_{ij}v_{ik}+ \\
    +\sum_{i=1}^n\frac{s_b}{s_xs_w}\left(\frac{s_b}{s_xs_w}-2d_i\right)v_i^2+2\frac{s_b}{s_xs_w}\sum_{i=1}^n\sum_{j=1}^f\tilde{x}_jv_iv_{ij}
\end{gathered}
\end{equation}
where $\tilde{x_j}$ is defined in Eq. \ref{eq: dequantization of variables} and 
\begin{equation}
    d_i=\frac{y_i}{s_ws_x}-\frac{s_b}{s_xs_w}\left\lfloor\frac{b_i}{s_ws_x}\right\rfloor- \sum_{j=1}^f \left\lfloor\frac{w_{ij}}{s_w}\right\rfloor{\text{round}\left(\frac{x_j}{s_x}\right)}.
\end{equation}
This structure allows it to be rewritten as a QUBO problem by organizing the binary variables into the vector $\mathbf{v}=(v_{11},...,v_{1f},...,v_{n1},...,v_{nf},v_1,...,v_n)$
and expressing the objective function as:
\begin{equation}
\mathcal{Q}(\mathbf{v}) = \mathbf{v}^{T} M \mathbf{v}.
\end{equation} 

Where $M$, defined in Eq. \ref{M_matrix}, is a coefficient matrix constructed from Eq.~\ref{B_term} that encodes all pairwise interactions between binary variables. The $M$ matrix is composed of 4 different types of sub-matrices ($M_i, T_i, T_i'$ and $D$). 

\begin{equation}\label{M_matrix}
M =
\left(
\begin{array}{c|c|c|c|c}
M_1 & 0 & \cdots & 0 & T_{i=1}' \\
\hline
0 & M_2 & \cdots & 0 & T_{i=2}' \\
\hline
\vdots & \vdots & \ddots & \vdots & \vdots \\
\hline
0 & 0 & \cdots & M_n & T_{i=n}' \\
\hline
T_{i=1} & T_{i=2} & \cdots & T_{i=n} & D
\end{array}
\right),
\end{equation}

There are $n$ matrices $M_i$, one for each value of $i$, these matrices are derived from the first and the second term of $\mathcal{Q}(\mathbf{v})$.

\begin{equation}
     M_{i}=\begin{pmatrix}
  \tilde{x}_1(\tilde{x}_1-2d_i) & \tilde{x}_1\tilde{x}_2 &  \tilde{x}_1\tilde{x}_3&\dotsi & \tilde{x}_1\tilde{x}_f\\
  \tilde{x}_1\tilde{x}_2 & \tilde{x}_2(\tilde{x}_2-2d_i) & \tilde{x}_2\tilde{x}_3 & \dotsi & \tilde{x}_2\tilde{x}_f\\
  \tilde{x}_1\tilde{x}_3 & \tilde{x}_2\tilde{x}_3 & \tilde{x}_3(\tilde{x}_3-2d_i) &\dotsi & \tilde{x}_3\tilde{x}_f\\
  \vdots & \vdots & \vdots & \ddots & \vdots \\
  \tilde{x}_1\tilde{x}_f & \tilde{x}_2\tilde{x}_f & \tilde{x}_3\tilde{x}_f & \dotsi & \tilde{x}_f(\tilde{x}_f-2d_i)
  \end{pmatrix}.
\end{equation}
The matrices $T_i$ and $T_i'$ are formed from the last term of $\mathcal{Q}(\mathbf{v})$ and only contain one non-zero row or column, respectively. 
\begin{equation}
     T_{i=1}=\frac{s_b}{s_xs_w}\begin{pmatrix}
  \tilde{x}_1 & \tilde{x}_2 &\dotsi & \tilde{x}_f\\
  0 & 0 & \dotsi & 0\\
  \vdots & \vdots & \vdots & \vdots\\
  0 & 0 & \dotsi & 0
  \end{pmatrix}; \quad T_{i=n}=\frac{s_b}{s_xs_w}\begin{pmatrix}
  0 & 0 & \dotsi & 0\\
  0 & 0 & \dotsi & 0 \\
  \vdots & \vdots & \vdots & \vdots\\
  \tilde{x}_1 & \tilde{x}_2 &\dotsi & \tilde{x}_f
  \end{pmatrix};
  \end{equation}
  
  \begin{equation}
     T_{i=1}'=\frac{s_b}{s_xs_w}\begin{pmatrix}
  \tilde{x}_1 & 0 &\dotsi & 0\\
  \tilde{x}_2 & 0 & \dotsi & 0\\
  \vdots & \vdots & \vdots & \vdots\\
  \tilde{x}_f & 0 & \dotsi & 0
  \end{pmatrix}; \quad      T_{i=n}'=\frac{s_b}{s_xs_w}\begin{pmatrix}
  0 & 0 & \dotsi & \tilde{x}_1 \\
 0 & 0 & \dotsi & \tilde{x}_2 \\
  \vdots & \vdots &  \vdots & \vdots\\
  0 & 0 & \dotsi & \tilde{x}_f
  \end{pmatrix}.
  \end{equation}\\
Finally, the last matrix is derived from the third term of $\mathcal{Q}(\mathbf{v})$
  \begin{equation}
     D=\frac{s_b}{s_xs_w}\begin{pmatrix}
  \frac{s_b}{s_xs_w}-d_1 & 0 &\dotsi & 0\\
  0 & \frac{s_b}{s_xs_w}-d_2 & \dotsi & 0\\
  \vdots & \vdots & \vdots & \vdots\\
  0& 0 & \dotsi & \frac{s_b}{s_xs_w}-d_n
  \end{pmatrix}.
  \end{equation}

When considering a dataset, instead of a single input-output pair, the error term is averaged across all samples:
\begin{equation}
    \mathbb{E}[\left\lVert\mathbf{y}-\check{\mathbf{y}}(\mathbf{v})\right\rVert^2]=(s_xs_w)^2\left(\mathbb{E}\left[\left\lVert\mathbf{y}-\underline{\mathbf{y}}\right\rVert^2\right]+\mathbb{E}[\mathcal{Q}(\mathbf{v})]\right),
\end{equation}
and the problem can be approximated and reformulated as the minimization of the cost function:
\begin{equation}\label{qubo_general}
    \mathbb{E}[\mathcal{Q}(\mathbf{v})]=\mathbf{v}^{T}\cdot{\mathbb{E}[M]}\cdot\mathbf{v}.
\end{equation}
Due to the symmetric structure of the $\mathbb{E}[M]$ matrix, it follows that the problem satisfies the criteria for a Quadratic Unconstrained Binary Optimization (QUBO) problem.
\section{Division of the problem into subproblems}\label{sec: subproblems}
The ADAROUND quantization problem has been defined as the process of determining the binary variables $v_{ij}$ and $v_i$, which decide whether each element of the weight matrix $w_{ij}$ and the bias $b_i$ should be quantized to the ceiling or the floor integer value. In order to solve this problem, it is needed to solve the QUBO problem of Eq.~\ref{qubo_general} associated with the matrix $M$ (Eq. \ref{M_matrix}). However, due to the high dimensionality and dense connectivity of this matrix, solving the full QUBO directly becomes computationally infeasible with conventional algorithms. For this reason, this paper proposes a methodology that allows simplifying the problem by dividing the problem into independent subproblems; this process is possible due to the symmetries of the $M$ matrix. 

Considering that $f$ denotes the dimensionality of $x$ and $n$ denotes the dimensionality of $y$, the associated QUBO problem involves $nf+n$ variables and a total of $nf^2+2nf+n$ variable interactions. However, by exploiting the block structure of $M$, the global problem can be decomposed into  $n$ QUBO subproblems. Each independent subproblem associated with a QUBO matrix $S_i$ is composed of the matrices $M_i$, the $i$-th row and column of the matrices $T_i$ and $T_i'$, and the $i$-th diagonal element of matrix $D$. 
\begin{equation}
    S_i =
\left(
\begin{array}{c|c}
M_i  & T_{i,[:,i]}' \\
\hline
T_{i,[i,:]}  & D_{ii}
\end{array}
\right),
\end{equation}
Therefore, the objective is to solve a set of QUBO subproblems of the form
\begin{equation}\label{eq: qubo_subproblem}
    \mathcal{Q}_i(\mathbf{v}_i) = \mathbf{v}_i^{\top} \mathbb{E}[S_i] \mathbf{v}_i,
\end{equation}

where each matrix $S_i$, defined in Equation~\ref{eq: sub_matrix}, is fully dense. Solving each subproblem provides the optimal configuration for the $i$-th set of binary variables, given by $\mathbf{v}_i = (v_{i1}, \dots, v_{if}, v_i)$.
These problems have $f+1$ variables and at most $f^2+2f+1$ interactions. Importantly, the independence of these subproblems enables Eq. \ref{qubo_general}  to be efficiently addressed by solving $n$ decoupled subproblems.
\begin{equation}\label{eq: sub_matrix}
     S_{i}=\begin{pmatrix}
  \tilde{x}_1(\tilde{x}_1-2d_i) & \tilde{x}_1\tilde{x}_2 &  \dotsi & \tilde{x}_1\tilde{x}_f & \frac{s_b}{s_xs_w}\tilde{x}_1\\
  \tilde{x}_1\tilde{x}_2 & \tilde{x}_2(\tilde{x}_2-2d_i)  & \dotsi & \tilde{x}_2\tilde{x}_f & \frac{s_b}{s_xs_w}\tilde{x}_2\\
\tilde{x}_1\tilde{x}_3 & \tilde{x}_2\tilde{x}_3  &\dotsi & \tilde{x}_3\tilde{x}_f & \frac{s_b}{s_xs_w}\tilde{x}_3\\
  \vdots & \vdots &  \ddots & \vdots & \vdots \\
  \tilde{x}_1\tilde{x}_f & \tilde{x}_2\tilde{x}_f  & \dotsi & \tilde{x}_f(\tilde{x}_f-2d_i) & \frac{s_b}{s_xs_w}\tilde{x}_f \\
  \frac{s_b}{s_xs_w}\tilde{x}_1 & \frac{s_b}{s_xs_w}\tilde{x}_2 & \dotsi & \frac{s_b}{s_xs_w}\tilde{x}_f & \frac{s_b}{s_xs_w}(\frac{s_b}{s_xs_w}-d_2) 
  \end{pmatrix}.
\end{equation}
\section{Algorithm implementation}\label{sec: implementation}
This section describes the steps required to implement QUBO-ADAROUND quantization of a neural network. The method extends the conventional PTQ round-to-nearest quantization by introducing the rounding criterion of each weight throughout a binary vector $\mathbf{v}$. For any integer precision level (int8, int4, ...), the algorithm can be summarized in the following steps:
\begin{enumerate}
    \item Computation of the scale points $s_w, s_b, s_x$ of the weights, bias and input of each layer using the quantization rules introduced in Section~\ref{subsec: quantization of variables} . 
    \item Construction of each matrix $\mathbb{E}[S_i]$ in Eq.~\ref{eq: sub_matrix} for a fixed number of inputs. It is important to note that only the diagonal terms differ between the matrices $\mathbb{E}[S_i]$ for each $i$.
    Each matrix is dense and contains all the interactions between the binary variables of $\mathbf{v}_i = (v_{i1}, \dots, v_{if}, v_i)$.
    
    \item Application of a generic QUBO solver for each subproblem $i$ to determine the optimal values of the binary variables $\mathbf{v}_i = (v_{i1}, \dots, v_{if}, v_i)$.
    \item Apply the resulting rounding variables to quantize the parameters of the neural network.
\end{enumerate}

There are numerous QUBO solvers that can be employed, including exact or approximate ones. As this methodology does not require the best possible solution, it is recommended to use a heuristic algorithm that obtains a sub-optimal solution very fast. In particular, the algorithm employed in the experimentation section is the D-Wave~\cite{dwave2025} implementation of the simulated annealing~\cite{simmulated_annealing_kirk}, a probabilistic metaheuristic, which iteratively explores the solution space to find
near-optimal solutions. The solver was executed with the default configuration parameters.
\section{Experimentation}\label{sec: experimentation}

\begin{table}[h!]
\centering
\label{tab:models}
\small
\begin{tabular}{|l|p{4.5cm}|p{6cm}|}
\hline
\textbf{Dataset} & \textbf{Layers (neurons)} & \textbf{Activation functions} \\ \hline
MNIST  & 1 layer: 10 & Softmax \\ \hline
MNIST  & 3 layers: 128, 64, 10 & ReLU, ReLU, Softmax \\ \hline
MNIST-FASHION  & 3 layers: 128, 64, 10 & ReLU, ReLU, Softmax \\ \hline
CIFAR-10 & 5 layers: 512, 256, 128, 64, 10 & ReLU, ReLU, ReLU, ReLU, Softmax \\ \hline
EMNIST   & 3 layers: 256, 128, 47 & ReLU, ReLU, Softmax \\ \hline
\end{tabular}
\caption{Summary of neural network architectures and activation functions used in the experiments. The size of the last layer is the number of different classes of each dataset.}
\label{table: neural network architectures}
\end{table}
\begin{table}[h!]
\centering
\resizebox{\textwidth}{!}{%
\begin{tabular}{|l|c|c|c|c|c|c|c|c|c|}
\hline
\textbf{Dataset} & \textbf{Original} & \multicolumn{2}{c|}{\textbf{int8}} & \multicolumn{2}{c|}{\textbf{int4}} & \multicolumn{2}{c|}{\textbf{int2}} & \multicolumn{2}{c|}{\textbf{int1}} \\
\hline
 &  & ADAR & RTN & ADAR & RTN & ADAR & RTN & ADAR & RTN \\
\hline
MNIST-1 & 0.9246 & 0.9246 & 0.9246 & \textbf{0.9219} & 0.9195 & \textbf{0.8968} & 0.6314 & \textbf{0.5267} & 0.1210 \\
\hline
MNIST-2 & 0.9804 & 0.9740& 0.9740 & \textbf{0.9720} & 0.9680 & \textbf{0.9360} & 0.6630 & 0.1010 & 0.1130       \\
\hline
MNIST-F & 0.8831 &  \textbf{0.8479} &  0.8467 & 0.8104 & \textbf{0.8324} & \textbf{0.5948} & 0.2868 & 0.1067 & 0.1067 \\
\hline
CIFAR-10 & 0.4185 & 0.2902  & \textbf{0.2913}  & \textbf{0.2889} & 0.2762 & \textbf{0.1816} & 0.1570 & 0.1000 & 0.1000\\
\hline
EMNIST & 0.8336 & \textbf{0.7082} & 0.7081  & \textbf{0.6728}  & 0.6717 & \textbf{0.2946} & 0.0543 &  0.0212 & 0.0212 \\
\hline

\end{tabular}
}
\caption{Classification accuracy for several datasets and network architectures under different quantization levels (int8, int4, int2, int1) using two quantization methods (ADAR: ADAROUND and RTN: round-to-nearest). 
\textbf{MNIST-1} corresponds to a single-layer network with 10 neurons; 
\textbf{MNIST-2} uses three layers with 128, 64, and 10 neurons; 
\textbf{MNIST-F} refers to the Fashion-MNIST dataset with layers of 128, 64 and 10 neurons; 
\textbf{CIFAR-10} is the grayscale version of the dataset with layers of 512, 256, 128, 64, and 10 neurons; 
\textbf{EMNIST} uses a three-layer model with 256, 128, and 47 neurons. The ADAROUND quantization is performed using a 10 $\%$ of the test data set. }
\label{table: dataset_results}
\end{table}

The QUBO-ADAROUND quantization was tested on four datasets: MNIST~\cite{mnist}, MNIST-FASHION~\cite{mnist-fashion}, EMNIST~\cite{emnist} and CIFAR-10~\cite{cifar-10}. For each dataset, a dense neural network with the architecture specified in Table~\ref{table: neural network architectures} was trained until convergence, and then the PTQ process was applied. The evaluation considers integer precisions ranging from int8 to int1, comparing the accuracy of the QUBO-ADAROUND method against the round-to-nearest. The methodology is implemented in Python, and the QUBO solver employed is simulated annealing, executed multiple times per subproblem, retaining the best solution.

The results are reported in Table~\ref{table: dataset_results}. At high-precision levels (int8 and int4), quantization is basically independent of the method. However, when aggressive quantization is employed (especially at int2), the ADAROUND method outperforms round-to-nearest across datasets, in some cases showing a very considerable difference. In the extreme case of int1, both methods perform poorly, indicating that dense neural networks cannot be compressed to binary precision without severe loss of expressivity.

\begin{figure}[h!]
    \centering
    % Figura 1
    \begin{minipage}[t]{0.48\linewidth}
        \centering
        \includegraphics[width=\linewidth]{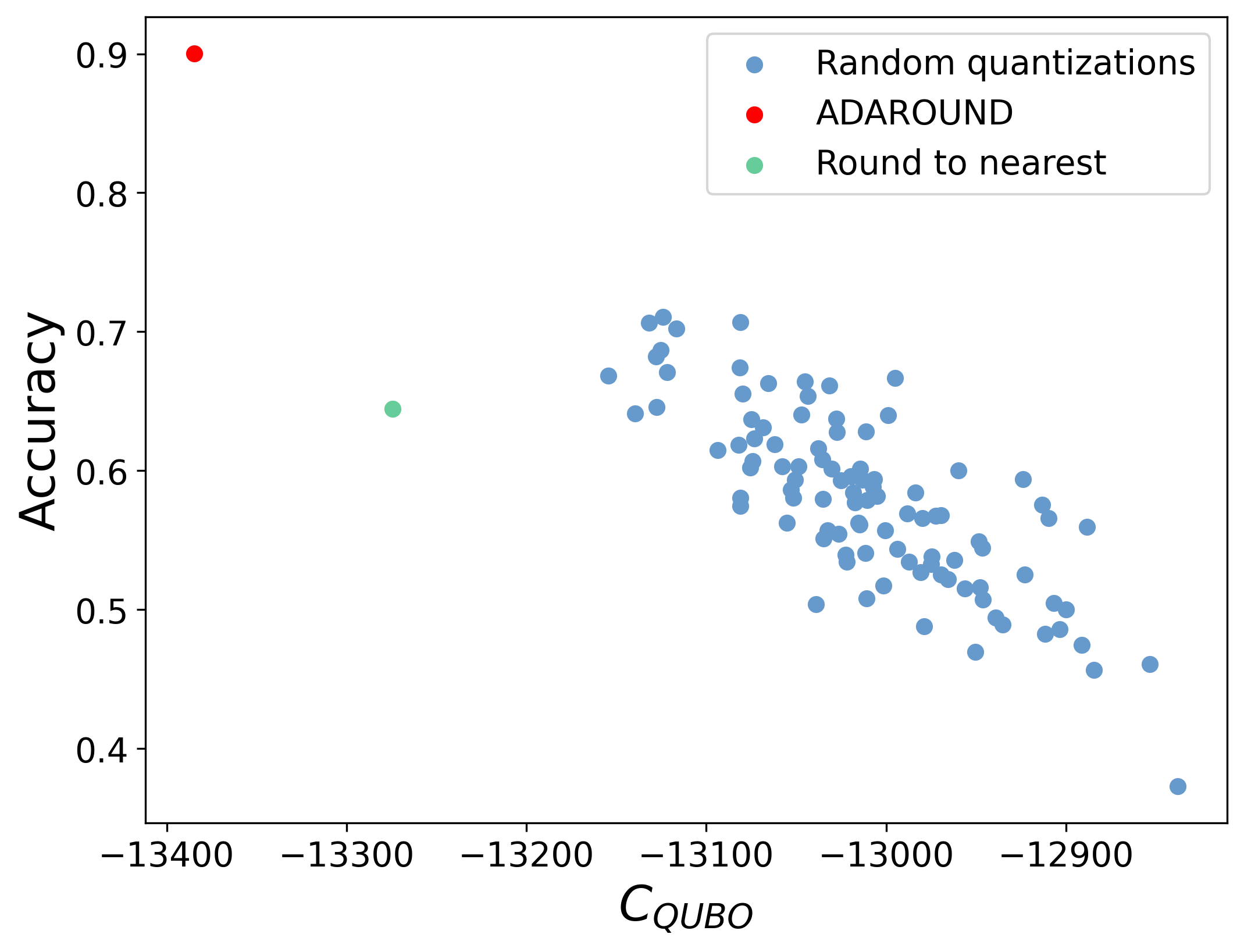}
        \caption{Accuracy vs $C_{QUBO} = \mathbf{v}^{T}\cdot{\mathbb{E}[M]}\cdot\mathbf{v}$ for a neural network of one layer with 10 neurons for the MNIST dataset quantized to int2. Random quantizations are shown in blue, ADAROUND in red, and rounding-to-the-nearest in green.}
        \label{fig:cost_mnist_int2}
    \end{minipage}
    \hfill
    % Figura 2
    \begin{minipage}[t]{0.48\linewidth}
        \centering
        \includegraphics[width=\linewidth]{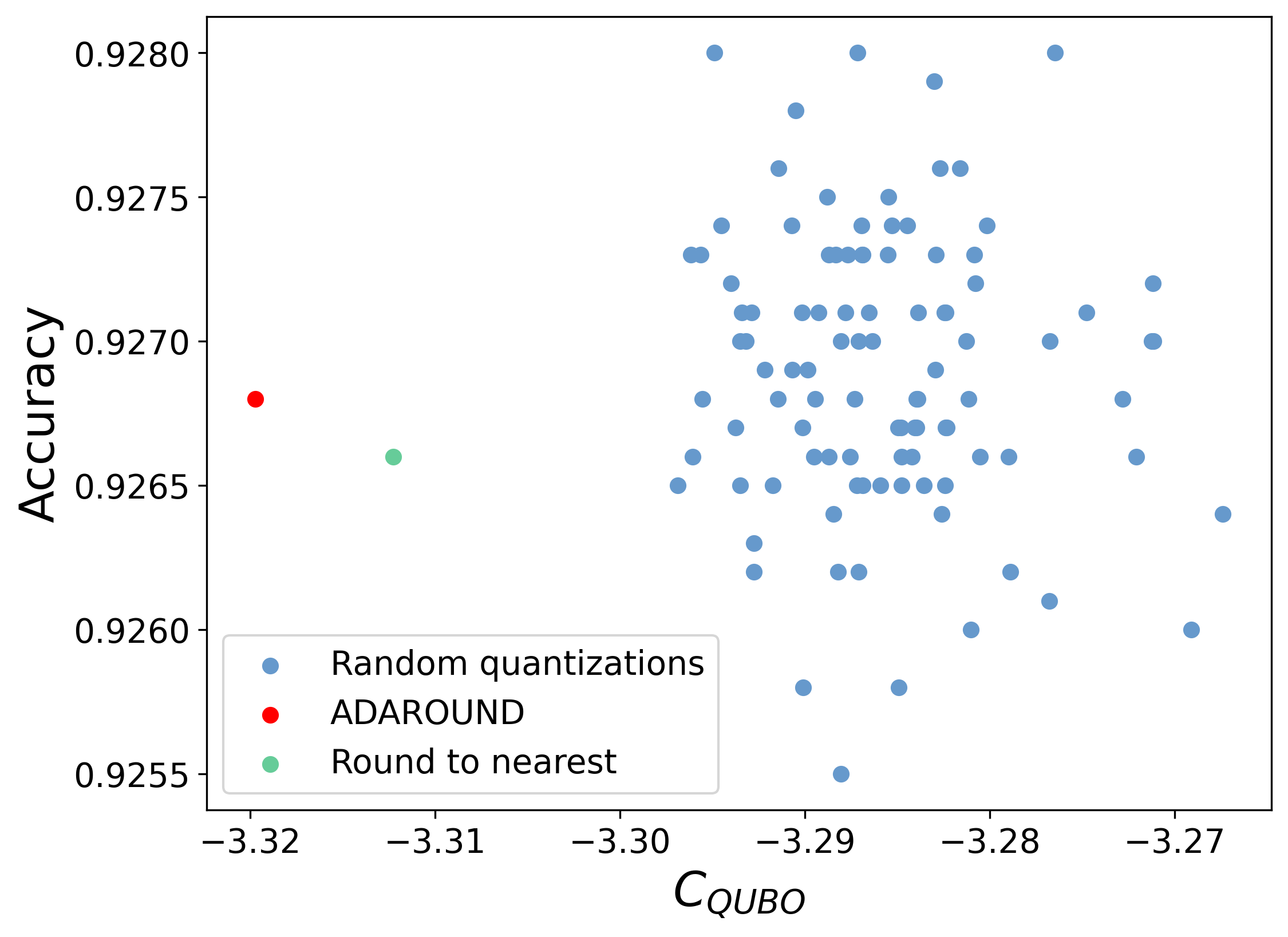}
        \caption{Accuracy vs $C_{QUBO} = \mathbf{v}^{T}\cdot{\mathbb{E}[M]}\cdot\mathbf{v}$ for a neural network of one layer with 10 neurons for the MNIST dataset quantized to int8. Random quantizations are shown in blue, ADAROUND in red, and rounding-to-the-nearest in green.}
        \label{fig:cost_mnist_int8}
    \end{minipage}
\end{figure}

An additional point of interest is the relationship between the network accuracy and the associated QUBO cost: $C_{QUBO} = \mathbf{v}^{T}\cdot{\mathbb{E}[M]}\cdot\mathbf{v}$. Optimizing the Frobenius difference between $\mathbf{y}$ and $\check{\mathbf{y}}$ does not necessarily lead to an improvement in the accuracy of the model. The exact nature of the relationship between these two metrics remains unclear. Fig.~\ref{fig:cost_mnist_int2} illustrates this dependence for the single-layer neural network on the MNIST dataset using int2 quantization. In this scenario, it is possible to appreciate the linear dependence between the magnitudes. Moreover, the ADAROUND quantization is significantly better than any randomly generated quantization. However, this dependence is not always satisfied for every neural network or for every type of quantization. For example, in the same model but quantized to int8, Fig.~\ref{fig:cost_mnist_int8} shows that there is no dependence between cost and accuracy.

Overall, these experiments show that the QUBO-ADAROUND method is particularly effective in such quantization levels, offering noticeable improvements over round-to-nearest in some situations. At the same time, they reveal its current limitations: performance collapses at int1 precision, and the method depends on heuristic solvers that do not guarantee optimality. Despite these constraints, the results confirm that the QUBO-ADAROUND represents a promising direction for PTQ, especially in scenarios where maintaining accuracy at very low bit-widths is critical. It should also be noted that the present experiments were limited to relatively small and medium-sized dense neural networks. The reason is that current QUBO solvers face scalability problems when dealing with very large problem instances.

\section{Conclusions}\label{sec: conclusions}
This paper introduces a different QUBO-based reformulation of the ADAROUND quantization for dense neural networks, aiming to minimize the Frobenius distance between the theoretical and the de-quantized output before the activation function. The main contribution of this work is the derivation of the mathematical formulation that expresses the quantization binary variable selection problem as a QUBO problem.  Additionally, due to the properties of the QUBO coefficient matrix $M$, it is proposed that it is possible to solve $n$ dense QUBO problems of $f+1$ variables, instead of a QUBO problem of $nf+n$ variables. In order to validate the methodology, the method was applied to the MNIST, Fashion-MNIST, EMNIST, and CIFAR-10 datasets. The results show that the quantization using the QUBO-ADAROUND performs significantly better than the round-to-nearest method in some precision levels, specifically at int2. Additionally, an empirical relationship has been observed between the network's accuracy and the value of the cost function $C_{QUBO}$. However, a satisfactory theoretical explanation for this behavior has not yet been established.

Future work could explore the application of this methodology to different datasets and alternative neural network architectures, such as LSTMs or LLMs. Another promising direction is the development of a more general version focused on multi-layer neural networks, or a formulation that enables quantization to a broader range of integer values rather than restricting it to the nearest integers. On a longer horizon, if practical quantum computing technologies become available, it would be worthwhile to investigate the use of quantum solvers to address the QUBO problems, potentially extending the scalability and applicability of the proposed methodology.

\section*{Acknowledgment}
The research has been funded by the Ministry of Science and Innovation and CDTI under ECOSISTEMAS DE INNOVACIÓN project ECO-20241017 (EIFEDE) and ICECyL (Junta de Castilla y León) under project CCTT5/23/BU/0002 (QUANTUMCRIP). This project has been funded by the Spanish Ministry of Science, Innovation and Universities under the project PID2023-149511OB-I00, and under the programme for mobility stays at foreign higher education and research institutions ``José Castillejo Junior" with code CAS23/00340.

\bibliographystyle{unsrtnat}
\bibliography{biblio}% Produces the bibliography via BibTeX.

\onecolumn\newpage
\appendix
\section{Mathematical derivation}\label{apendix: mathematical_dev}
This appendix section shows the mathematical calculation of the Frobenius distance $\left\lVert\mathbf{y}-\check{\mathbf{y}}(\mathbf{v})\right\rVert^2$ between the theoretical output $\mathbf{y}$ and the dequantized output $\check{\mathbf{y}}(\mathbf{v})$, both before activation. In Eq. \ref{eq: output dequantized}, the dequantized output has been defined as 
\begin{equation*}
\check{y_i}(\mathbf{v})=\sum_{j=1}^f s_ws_x\underline{w_{ij}}\tilde{x}_j+s_b\underline{b_i} + \sum_{j=1}^f s_ws_xv_{ij}\tilde{x}_j+s_bv_i.
\end{equation*}
This equation can be reformulated by multiplying and dividing by $s_xs_w$ to the terms that are multiplied by $s_b$
\begin{equation}
\begin{gathered}
    \check{y_i}(\mathbf{v})= \sum_{j=1}^f s_ws_x\underline{w_{ij}}\tilde{x}_j + \frac{s_bs_xs_w}{s_xs_w}\underline{b_i} +
    \sum_{j=1}^f s_ws_xv_{ij}\tilde{x}_j + \frac{s_bs_xs_w}{s_xs_w}v_i=\\
    =s_ws_x \sum_{j=1}^f \left [\underline{w_{ij}}\tilde{x}_j + \frac{s_b}{s_xs_w}\underline{b_i}\right] + s_ws_x \sum_{j=1}^f\left [ v_{ij}\tilde{x}_j + \frac{s_b}{s_x s_w}v_i \right],
    \end{gathered}
\end{equation}
In order to simplify the notation, it is possible to rewrite the expression using $\underline{b_i}'=\frac{s_b}{s_xs_w}\underline{b_i}$, $v_i'=\frac{s_b}{s_xs_w}v_i$ and $\underline{y_i}=\sum_{j=1}^f\underline{w_{ij}}\tilde{x}_j+\underline{b_i}'$:
\begin{equation}
    \check{y_i}(\mathbf{v})=s_ws_x\left(\underline{y_i}+ \sum_{j=1}^f v_{ij}\tilde{x}_j+v_i'\right).
\end{equation}
Using this expression, the Frobenius distance can be expanded more conveniently.
\begin{equation}
    \begin{gathered}
    \left\lVert\mathbf{y}-\check{\mathbf{y}}(\mathbf{v})\right\rVert^2=\sum_{i=1}^n(y_i-\check{y_i})^2= \sum_{i=1}^n \left [ y_i^2+\check{y_i}(\mathbf{v})^2-2y_i\check{y}_i(\mathbf{v}) \right ]=\\
    = (s_w s_x)^2 \sum_{i=1}^n \left [ \frac{y_i^2}{(s_w s_x)^2} + 
\left( \underline{y_i}+ \sum_{j=1}^f v_{ij}\tilde{x}_j+v_i'\right)^2
- 2 \frac{y_i}{s_w s_x} 
\left( \underline{y_i}+ \sum_{j=1}^f v_{ij}\tilde{x}_j+v_i \right) 
\right ].
    \end{gathered}
\end{equation}
By performing the change of variable $y_i=\frac{y_i}{s_ws_x}$ and expanding:
\begin{equation}
\begin{gathered}
    \frac{\left\lVert\mathbf{y}-\check{\mathbf{y}}(\mathbf{v})\right\rVert^2}{(s_w s_x)^2} =
    \sum_{i=1}^n  \left[y_i^2 + \underline{y_i}^2 
+ 2 \underline{y_i} \sum_{j=1}^fv_{ij} \tilde{x}_j + 2 \, \underline{y_i} v'_{i} 
+ \sum_{j,k}^{f,f}v_{ij} v_{ik} \tilde{x}_j \tilde{x}_k 
+\right. \\
\left. 2 v'_i \sum_{j=1}^f v_{ij}  \tilde{x}_j + {v_i'}^2
- 2 y_i \underline{y_i} 
- 2 y_i \sum_{j=1}^f v_{ij} \tilde{x}_j 
- 2 y_i v'_{i} \right].
\end{gathered}
\end{equation}
The Frobenius distance between $\mathbf{y}$ and $\underline{\mathbf{y}}$ is included in the previous equation, so it can be reformulated. Doing this, it is possible to have a term that depends on $\mathbf{v}$ and a term that does not.
\begin{equation}
\begin{gathered}
    \frac{\left\lVert\mathbf{y}-\check{\mathbf{y}}(\mathbf{v})\right\rVert^2}{(s_w s_x)^2} = \left\lVert\mathbf{y}-\underline{\mathbf{y}}\right\rVert^2 + 
    \sum_{i=1}^n \left[2 \underline{y_i} \sum_{j=1}^fv_{ij} \tilde{x}_j + 2 \, \underline{y_i} v'_{i} 
+ \sum_{j,k}^{f,f}v_{ij} v_{ik} \tilde{x}_j \tilde{x}_k 
+\right. \\
\left. 2 v'_i \sum_{j=1}^f v_{ij}  \tilde{x}_j + {v_i'}^2
- 2 y_i \sum_{j=1}^f v_{ij} \tilde{x}_j 
- 2 y_i v'_{i} \right].
\end{gathered}
\end{equation}
In order to clarify the expression, it is possible to rename \( d_i = y_i - \underline{y_i} \)

\begin{equation}
\begin{gathered}
    \frac{\left\lVert\mathbf{y}-\check{\mathbf{y}}(\mathbf{v})\right\rVert^2}{(s_w s_x)^2} = \left\lVert\mathbf{y}-\underline{\mathbf{y}}\right\rVert^2 + 
    \sum_{i=1}^n \biggr [-2 d_i \sum_{j=1}^fv_{ij} \tilde{x}_j - 2 \, d_i v'_{i} 
+ \sum_{j,k}^{f,f}v_{ij} v_{ik} \tilde{x}_j \tilde{x}_k 
+ \\
2 v'_i \sum_{j=1}^f v_{ij}  \tilde{x}_j + {v_i'}^2
 \biggl].
\end{gathered}
\end{equation}

Reaching the expression of the Frobenius distance $\left\lVert\mathbf{y}-\check{\mathbf{y}}(\mathbf{v})\right\rVert^2$. It is important to note that this result applies only to a single input vector $\mathbf{x}$ and its output vector $\mathbf{y}$. If one is interested in computing the error metric on a dataset $\{(\mathbf{x}_{\text{in}}^\alpha,\mathbf{y}_\text{out}^\alpha)\}_{\alpha=1,...,t}$, then the average over the different metrics in the previous expression must be taken, and the task would be to minimize that average:
\begin{equation}
\begin{gathered}
    \mathbb{E}\left[\left\lVert\mathbf{y}-\check{\mathbf{y}}(\mathbf{v})\right\rVert^2\right] = \frac{(s_w s_x)^2}{t} \sum_{\alpha=1}^t \Biggl[ 
\left\lVert\mathbf{y}^\alpha-\underline{\mathbf{y}^\alpha}\right\rVert^2 + 
    \sum_{i=1}^n \biggr [-2 d_i \sum_{j=1}^fv_{ij} \tilde{x}_j^\alpha - 2 \, d_i v'_{i} \\
+ \sum_{j,k}^{f,f}v_{ij} v_{ik} \tilde{x}_j^\alpha \tilde{x}_k^\alpha + 2 v'_i \sum_{j=1}^f v_{ij}  \tilde{x}_j^\alpha + {v_i'}^2
 \biggl]
    \Biggr]
    \end{gathered}
\end{equation}
To simplify the development, this expression will be set aside for now, but it should be kept in mind because, in real-world problems, the minimization of the error metric must be carried out over a dataset.

It is possible to expand $\left\lVert\mathbf{y}-\check{\mathbf{y}}(\mathbf{v})\right\rVert^2$ to have a purely QUBO expression. Taking into account that $v_{ij} = v_{ij}^2$ are binary variables
and that the crossed term can be expressed as:

\begin{equation}
\sum_{i,j,k}^{n,f,f}v_{ij} v_k^i \tilde{x}^j \tilde{x}^k = 
        \sum_{i=1}^n \sum_{j=1}^f \tilde{x}_j^2 v_{ij}^2 
        + \sum_{i=1}^n \sum_{j=1}^f \sum_{\substack{k=1 \\ k \neq j}}^f \tilde{x}_j \tilde{x}_k v_{ij} v_{ik}
\end{equation}

Then, the Frobenius distance can be expressed as follows:
\begin{equation*}
    \left\lVert\mathbf{y}-\check{\mathbf{y}}(\mathbf{v})\right\rVert^2=(s_ws_x)^2\left[\left\lVert\mathbf{y}-\underline{\mathbf{y}}\right\rVert^2+\mathcal{Q}(\mathbf{v})\right],
\end{equation*}
\begin{equation*}
\begin{gathered}
\mathcal{Q}(\mathbf{v})=\sum_{i=1}^n\sum_{j=1}^f\tilde{x}_j(\tilde{x}_j-2d_i)v_{ij}^2+\sum_{i=1}^n\sum_{j=1}^f\sum_{k=1, k{\neq}j}^f\tilde{x}_j\tilde{x}_kv_{ij}v_{ik}+ \\
    +\sum_{i=1}^n\frac{s_b}{s_xs_w}\left(\frac{s_b}{s_xs_w}-2d_i\right)v_i^2+2\frac{s_b}{s_xs_w}\sum_{i=1}^n\sum_{j=1}^f\tilde{x}_jv_iv_{ij}.
\end{gathered}
\end{equation*}

\end{document}